\documentclass[11pt]{article}

\usepackage[margin=0.8in]{geometry}
\usepackage[numbers]{natbib}
\usepackage{parskip}
\usepackage{authblk}
\usepackage{graphicx}
\usepackage{amsmath,amssymb,amsfonts}
\usepackage{url}
\usepackage[hidelinks]{hyperref}
\usepackage{xcolor}

\title{Scalable Evaluation of the Realism of Synthetic Environmental Augmentations in Images}

\author[1]{Damian Ruck}
\author[1]{Paul Vautravers}
\author[1]{Oliver Chalkley}
\author[1]{Jake Thomas}
\affil[1]{Advai Ltd, 20--22 Wenlock Road, London, N1 7GU, United Kingdom }

\date{} % no date

\begin{document}

\maketitle
\begin{abstract}
Evaluation of AI systems often requires synthetic test cases, particularly for rare or safety-critical conditions that are difficult to observe in operational data. Generative AI offers a promising approach for producing such data through controllable image editing, but its usefulness depends on whether the resulting images are sufficiently realistic to support meaningful evaluation.

We present a scalable framework for assessing the realism of synthetic image-editing methods and apply it to the task of adding environmental conditions---fog, rain, snow, and nighttime---to car-mounted camera images. Using 40 clear-day images, we compare rule-based augmentation libraries with generative AI image-editing models. Realism is evaluated using two complementary automated metrics: a vision-language model (VLM) jury for perceptual realism assessment, and embedding-based distributional analysis to measure similarity to genuine adverse-condition imagery.

Generative AI methods substantially outperform rule-based approaches, with the best generative method achieving approximately 3.6 times the acceptance rate of the best rule-based method. Performance varies across conditions: fog proves easiest to simulate, while nighttime transformations remain challenging. Notably, the VLM jury assigns imperfect acceptance even to real adverse-condition imagery, establishing practical ceilings against which synthetic methods can be judged. By this standard, leading generative methods match or exceed real-image performance for most conditions.

These results suggest that modern generative image-editing models can enable scalable generation of realistic adverse-condition imagery for evaluation pipelines. Our framework therefore provides a practical approach for scalable realism evaluation, though validation against human studies remains an important direction for future work.

\end{abstract}

\clearpage

\section{Introduction}

Reliable evaluation of AI systems destined for real-world deployment demands methods that scale to rare and safety-critical conditions. Two pressures make this need particularly acute. First, in safety-critical applications such as healthcare and autonomous driving, pre-deployment evaluation must often proceed before meaningful operational data can be collected---waiting to encounter rare failure modes in the field is unacceptable. Second, even where operational data does exist in quantity, the safety-critical edge cases that matter most---unusual weather, rare lighting conditions, unexpected object configurations---remain vanishingly rare. Together, these pressures create a persistent need for the targeted generation of synthetic test cases that cover the conditions an AI system must handle.

Generative AI is a compelling tool for meeting this need. Modern foundation models accept free-form natural-language instructions, producing outputs that increasingly approximate the richness and variability of human-created content. This matters because it is usually humans---domain experts, annotators, test engineers---on whom we rely to design, curate, and produce the data against which AI systems are evaluated~\cite{summerfield2025strange}. A generative model that can follow a prompt such as ``add heavy rain with wet road surfaces to this image'' offers the prospect of automating much of that human role, enabling evaluation data to be produced at scale, on demand, and with fine-grained control over the conditions depicted. The diversity of generative image editing approaches has been extensively taxonomized in recent surveys~\cite{huang2025diffusion}, covering training-based methods, finetuning approaches, and training-free techniques across numerous editing paradigms.

Controllability alone, however, is not sufficient. In domains such as imagery, synthetic data must also be \textit{realistic}: if the generated conditions do not faithfully represent the real-world phenomena they are meant to simulate, then any evaluation performed on that data tells us little about how the AI system would behave under genuine operational conditions. This challenge is well documented as the \textit{simulation-to-reality gap}~\cite{song2024synthetic, tobin2017domain}. Although it has long been possible to generate synthetic data for training and testing AI systems---whether through physics engines rendering three-dimensional driving scenes or simpler procedural environments for reinforcement learning---a persistent gap has existed between synthetic and real-world distributions. Behaviour observed under synthetic conditions does not necessarily predict behaviour under real conditions, limiting the trustworthiness of conclusions drawn from synthetic evaluation data.

The severity of the simulation-to-reality gap varies across generation methods and conditions. Simple transformations such as contrast reduction may approximate fog adequately, while more complex transformations—such as day-to-night conversion—require global scene changes that simple techniques cannot capture. Consequently, the realism of synthetic evaluation data must be assessed on a case-by-case basis. A scalable and automated evaluation procedure is therefore essential for practitioners who wish to build and maintain reliable evaluation datasets.

In this paper, we present a scalable framework for evaluating the realism of environmental augmentations in car-mounted imagery, and benchmark rule-based and generative editing approaches across four adverse conditions. Using clear-day images from the Adverse Conditions Dataset with Correspondences (ACDC)~\cite{sakaridis2021acdc}, we systematically compare two paradigms---traditional rule-based augmentation libraries (imgaug, albumentations) and modern generative AI image-editing models (OpenAI GPT-Image-1, Google Gemini, Alibaba Qwen, Flux Kontext)---across four target conditions: fog, rain, snow, and nighttime.

We evaluate realism using a VLM jury and an embedding-based distributional analysis. By seeking convergence between these independent evaluation modalities, we build justified confidence in comparative conclusions without requiring expensive human annotation at every iteration. Prior work~\cite{huang2025diffusion} has shown that multimodal model-based evaluation can correlate strongly with human judgments in image editing tasks, although such results are typically demonstrated in general editing benchmarks rather than domain-specific settings. Consequently, validation against human judgments remains an important direction for future work in the context of environmental augmentation.

Throughout this paper, we define \textit{realism} as the degree to which a synthetically augmented image is perceptually and statistically indistinguishable from a genuine image captured under the target environmental condition. This encompasses both the convincingness of the depicted condition (e.g., does the rain look like real rain?) and the preservation of scene semantics (e.g., are objects and spatial relationships maintained?). We do not address \textit{control}---the reliability with which a method produces data matching a specified target---which we leave for future work.

\section{Related Work}

\subsection{Physics-based and Traditional Augmentation}

Early adverse-condition augmentation relied on rule-based simulation using physical models. Sakaridis et al.\ introduced Foggy Cityscapes~\cite{sakaridis2018semantic}, applying Beer--Lambert attenuation with depth-based fog rendering to Cityscapes images. Training on these augmentations improved semantic segmentation on real foggy scenes. Von Bernuth et al.~\cite{vonbernuth2019simulating} extended this work to snow and fog through three-dimensional scene reconstruction and OpenGL rendering with physically motivated parameters, showing synthetic weather degraded object detection performance similar to real conditions.

Tremblay et al.~\cite{tremblay2021rain} developed rain rendering techniques combining physics-based particle simulation, GAN-based rain streak synthesis, and hybrid methods. User studies rated their hybrid approach more realistic than previous methods. Training on rain-augmented data yielded notable improvements in detection and segmentation on real rainy scenes.

\subsection{Generative Adversarial Networks and Style Transfer}

Generative adversarial networks (GANs) enabled learning environmental transformations directly from data. Multiple studies applied CycleGAN~\cite{zhu2017unpaired} to BDD100K~\cite{yu2020bdd100k} data for unpaired translation between weather domains and day--night conditions, capturing complex stylistic elements such as road reflections and headlight glow. Models trained with GAN-augmented data demonstrated improved generalization.

Mukherjee et al.~\cite{mukherjee2022generative} proposed attribute-conditioned GANs enabling fine-grained control over weather and illumination. Models trained with these augmentations showed better generalization and increased robustness. Gupta et al.~\cite{gupta2024robust} directly compared analytical, GAN-based, and style transfer augmentation for BDD100K object detection, finding style transfer yielded the highest performance on real adverse-condition tests, balancing neural realism with structural preservation better than pure GAN approaches.

\subsection{Diffusion Models and Large-Scale Generative Systems}

Recent work leverages diffusion models and text-to-image systems for environmental augmentation. Gurbindo et al.~\cite{gurbindo2025object} employed InstructPix2Pix~\cite{Brooks_2023_CVPR} to add weather effects via natural language prompts, with edited images retaining scene semantics while introducing target condition characteristics. Rothmeier et al.~\cite{rothmeier2024time} generated large numbers of synthetic driving scenes in adverse conditions using Midjourney, reporting detection gains on real adverse-condition datasets. Their fully synthetic generative data outperformed simple overlay techniques that provided minimal benefit or, in some cases, degraded performance.

Assion et al.~\cite{assion2024abdd} developed A-BDD, augmenting BDD100K with synthetic adverse weather and lighting conditions. They evaluated augmentation quality through downstream detection performance rather than direct realism assessment, finding that properly calibrated synthetic augmentations improved model robustness on held-out test sets.

\subsection{Automated Evaluation of Image Editing Methods}

Huang et al.~\cite{huang2025diffusion} provide an exhaustive survey of diffusion-based image editing, categorizing over 100 methods and introducing EditEval for evaluating general text-guided editing tasks. While such taxonomies establish the breadth of generative editing capabilities, our work addresses a complementary gap: systematic realism evaluation for domain-specific environmental augmentation in safety-critical applications. Unlike EditEval's assessment of editing accuracy, contextual preservation, and visual quality across diverse tasks, we evaluate distributional similarity and perceptual indistinguishability from genuine adverse-condition imagery—properties that are essential when synthetic data validates autonomous driving systems.

Their study also reports empirical analysis of automated evaluation metrics for image editing tasks. In the EditEval benchmark, scores produced by multimodal language models exhibit strong correlation with human study ratings across several editing dimensions, with Pearson correlations typically ranging from approximately 0.72 to 0.94 depending on editing category. These correlations substantially exceed those achieved by CLIP-based similarity metrics and other automated measures. While this evidence suggests that multimodal model-based evaluation can approximate human perceptual judgments of editing realism, the evaluation focuses on general text-guided editing tasks rather than domain-specific environmental augmentation. Consequently, the correlations reported in Huang et al.\ should not be assumed to transfer directly to the evaluation setting considered here.

\subsection{Gaps and Positioning}

Despite comprehensive surveys of diffusion-based editing techniques~\cite{huang2025diffusion} and growing work on adverse-condition augmentation, critical gaps remain for practitioners building evaluation datasets for safety-critical systems. Most studies evaluate augmentation through single validation modalities---either downstream task performance or human studies---rather than integrated frameworks combining multiple realism metrics. Comprehensive benchmarks systematically comparing traditional rule-based and foundation model techniques under identical conditions are limited. Embedding-based distributional metrics receive insufficient attention as standalone validation tools, typically serving as filters rather than primary realism indicators. Further, VLM juries (or ensembles of VLM judges) are under-used in evaluating properties of images in the evaluation of autonomous systems. 

This work addresses these gaps through a comprehensive evaluation framework combining VLM-Jury and embedding space distributional analysis. We systematically compare rule-based augmentation libraries against AI image editing models on identical clear-day images, establishing controlled conditions enabling direct comparison of realism. By evaluating these synthetic images against genuine adverse-condition imagery from ACDC, we also quantify both perceptual realism and statistical similarity to a real-world baseline, providing practitioners with actionable guidance for selecting augmentation methods suitable for safety-critical evaluation datasets.

\section{Methodology}

\begin{figure}[htbp]
    \centering
    \includegraphics[width=0.95\linewidth]{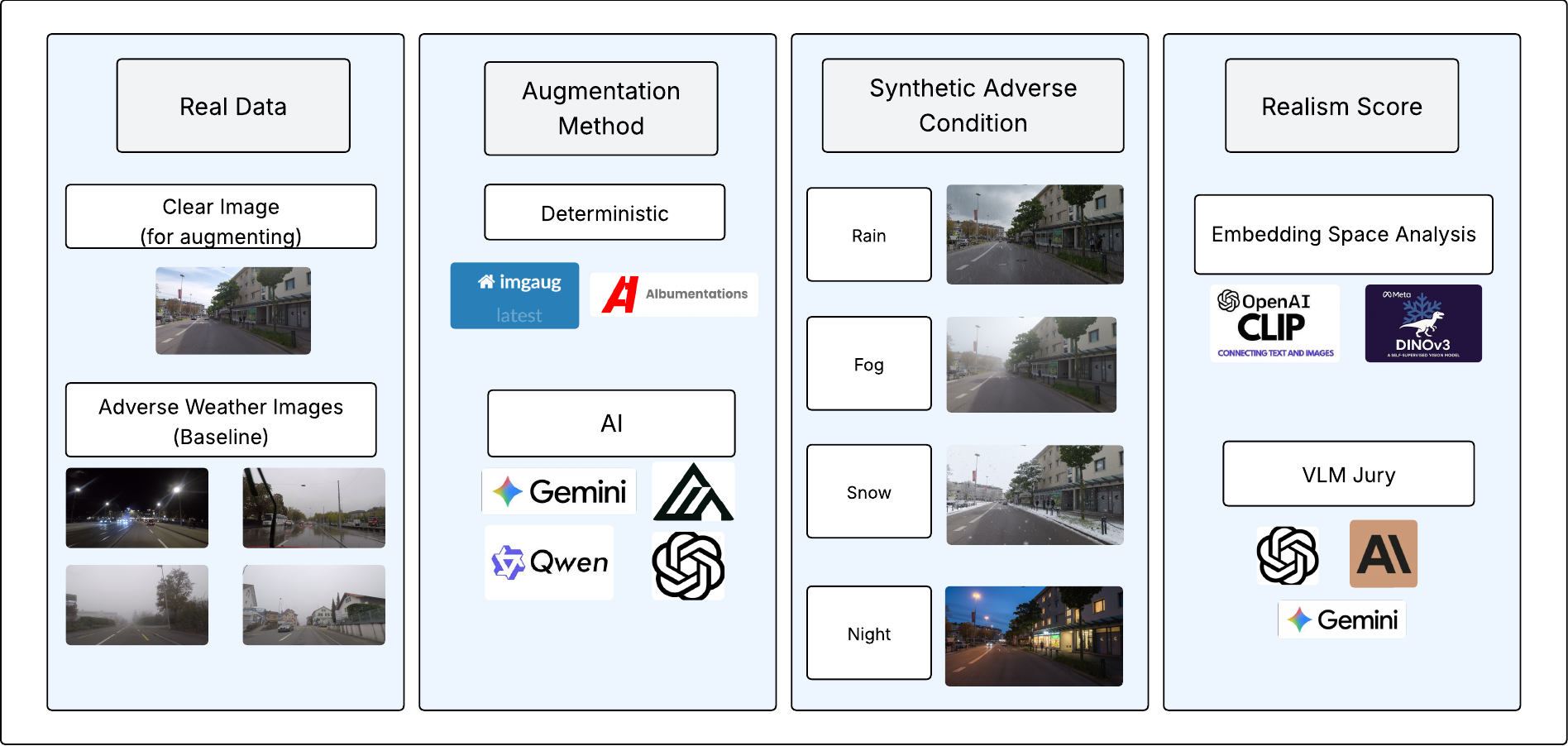}
    \caption{Overview of the experimental framework. Clear-day images are transformed using rule-based or generative AI methods to simulate adverse environmental conditions. Realism is assessed via VLM jury evaluation and embedding-based distributional analysis using CLIP and DINOv3.}
    \label{fig:method}
\end{figure}

This study evaluates the realism of synthetic environmental augmentation techniques applied to autonomous driving imagery. We compare two paradigms:  rule-based transformations and generative AI image editing models. Our evaluation employs two complementary assessment methods: a VLM jury that judges perceptual realism and preservation of semantic features in the original image, and an embedding-based distributional analysis that measures proximity to real adverse-condition imagery in foundation model feature spaces. Figure~\ref{fig:method} provides an overview of the experimental framework.

\subsection{Dataset}

We source imagery from the ACDC (Adverse Conditions Dataset with Correspondences) dataset \cite{sakaridis2021acdc}, which contains driving scenes captured across diverse weather and lighting conditions including fog, rain, snow, and nighttime conditions. The dataset also provides real-world examples of each environmental condition that serve as a baseline for assessing each augmentation method.

From the ACDC dataset, we randomly sample 40 clear-day images to serve as our test set for synthetic augmentation. These source images depict typical daytime driving scenarios without adverse conditions, providing a consistent baseline from which all augmentation methods can be applied. The remaining 3,566 real adverse-condition images (excluding these 40 test images) are reserved for constructing reference distributions in our embedding-based evaluation.

\subsection{Augmentation Methods}

We evaluate six augmentation methods spanning both rule-based and generative paradigms. All methods transform the same 40 source images to simulate four target conditions: rain, snow, fog, and nighttime.

\subsubsection{Rule-based Methods}

We employ two widely-used rule-based augmentation libraries: \texttt{imgaug} \cite{imgaug2020} and \texttt{albumentations} \cite{buslaev2020albumentations}. Both rely on hand-crafted heuristics and predefined parameter ranges rather than learned representations of environmental phenomena. Augmentations are achieved through combinations of operations such as additive Gaussian noise for precipitation, brightness reduction for fog, and gamma adjustments for nighttime conditions.

While computationally efficient and fully rule-based, these methods cannot capture the complex, context-dependent visual characteristics of real adverse conditions---such as reflections on wet surfaces, physically plausible snow accumulation patterns, or realistic atmospheric scattering in fog.

\subsubsection{Generative AI Methods}

We evaluate four AI model-based image editing systems: OpenAI's GPT-Image-1 \cite{openai2025gptimage}, Google's Gemini 2.5 Flash \cite{google2025gemini}, Alibaba's Qwen Image Edit Plus \cite{qwen2025imageedit}, and Black Forest Labs' Flux Kontext \cite{blackforest2025flux}.

Each model receives free-form natural language prompts specifying the desired environmental transformation (e.g., ``Transform this image to show heavy rain with wet road surfaces''). Images are processed in batches of 5, with each model receiving semantically equivalent prompts for each environmental condition. This ensures that performance differences reflect model capabilities rather than prompt engineering variations. Full prompts are provided in Appendix~\ref{app:prompts}.

Figure~\ref{fig:examples} shows representative outputs from each augmentation method alongside real adverse-condition imagery.

\begin{figure}[htbp]
    \centering
    \includegraphics[width=0.85\linewidth]{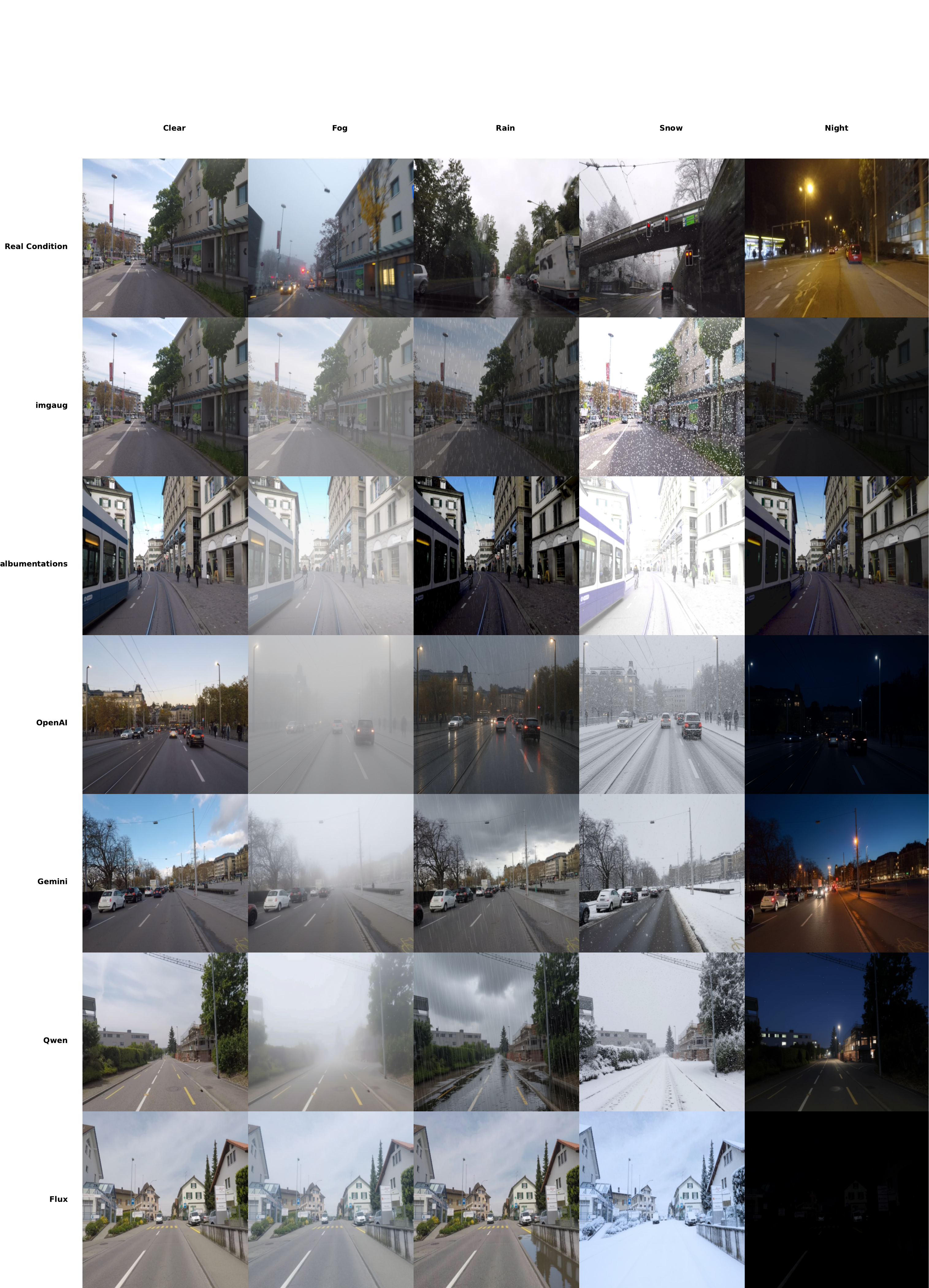}
    \caption{Example augmentations across methods and target conditions. Top row shows real ACDC images. Subsequent rows show outputs from Rule-based Methods (imgaug, albumentations) and generative AI methods (OpenAI, Gemini, Qwen, Flux).}
    \label{fig:examples}
\end{figure}

\subsection{VLM Jury Evaluation}

Our primary evaluation uses an ensemble of three VLMs as judges: GPT-4o \cite{openai2024gpt4o} (OpenAI), Claude Sonnet 4 \cite{anthropic2025claude} (Anthropic), and Gemini 2.5 Pro \cite{google2025gemini} (Google). This jury-based approach mitigates individual model biases and provides a more robust assessment of realism than relying on any single model.

We additionally perform post-hoc analysis of failure reasons to distinguish semantic preservation failures from realism failures; methodology is detailed in Appendix~\ref{app:failure_analysis}.

We employ binary accept/reject decisions rather than scalar ratings for several methodological reasons grounded in recent empirical findings. First, Large Language Models (LLMs) exhibit well-documented calibration problems when assigning absolute scores on Likert scales, including scale collapse where models cluster around specific values and round-number bias that reduces discriminative power~\cite{liu2023geval}. Second, recent work directly comparing evaluation protocols found that absolute scoring is more robust to manipulation than pairwise comparison: pairwise preferences flip in approximately 35\% of cases when distractor features are introduced, compared to only 9\% for absolute scores~\cite{tripathi2025pairwise}. For objective quality assessment tasks such as realism verification, pointwise approaches provide a more reliable measure by design, as binary preferences can exaggerate small quality gaps~\cite{tripathi2025pairwise}. Third, pairwise evaluation suffers from well-documented position bias, verbosity bias, and self-enhancement bias that complicate interpretation~\cite{zheng2023judging, ye2025justice}. While pairwise comparisons can better detect subtle quality differences in non-adversarial conditions~\cite{jeong2024prepair}, our evaluation context---where edge-case and potentially ambiguous samples are common---favours the robustness of binary pointwise decisions. By aggregating many binary decisions across multiple independent judges, we recover continuous acceptance rates with principled confidence intervals while avoiding the calibration challenges inherent in multi-level scalar ratings.

\subsubsection{Evaluation Protocol}

Each judge evaluates augmented images by examining the original clear-day image alongside its transformed counterpart as an image pair. Judges assess two criteria.

First, condition realism: does the transformed image convincingly depict the target condition? For each condition, judges receive specific guidance. Snow augmentations should exhibit evidence of falling or accumulated snow. Rain augmentations should show precipitation or wet surfaces. Fog augmentations should demonstrate atmospheric obscuration with reduced visibility. Night transformations should feature appropriate darkness, artificial lighting, and illumination characteristics.

Second, semantic preservation: apart from condition-appropriate changes, are the scene semantics preserved? Objects, spatial relationships, and scene structure should remain consistent with the original image, with exceptions only for elements that would naturally be obscured or altered by the target condition (e.g., reduced visibility of distant objects in fog).

Each judge returns a binary decision: \textit{true} if both criteria are satisfied, \textit{false} otherwise. We aggregate decisions across all three judges to compute mean acceptance rates per method and condition.

\subsubsection{Baseline Calibration}

To calibrate the VLM jury's standards and establish upper bounds on expected performance, we evaluate genuine adverse-condition imagery using the same judges. We sample 40 real images per adverse environmental condition from the ACDC dataset and present these to each judge individually. Each judge evaluates whether the image convincingly depicts its labelled condition, using condition-specific criteria (e.g., ``Does this image convincingly look like it was taken during rain?'').

This baseline reveals the judges' inherent acceptance rates for authentic adverse-condition imagery. It provides essential context for interpreting synthetic augmentation scores: if real rain images achieve only 93.3\% acceptance, then a synthetic method achieving 93\% is performing at near-ceiling levels. The baseline also exposes potential judge biases and calibration differences. Full baseline results and inter-judge agreement analysis are provided in Appendix~\ref{app:vlm_reliability}.

\subsection{Embedding-based Distributional Analysis}

Our second evaluation method assesses whether augmented images occupy similar regions of semantic feature space as real adverse-condition images. This provides a realism assessment that is largely independent of the VLM jury approach, using different models and evaluation criteria to build confidence through convergent evidence.

\subsubsection{Image Embedding Models}

We extract semantic representations using two pre-trained vision encoders that capture different aspects of visual content.

CLIP (Contrastive Language-Image Pre-training) \cite{radford2021clip} from OpenAI produces embeddings aligned with natural language semantics through contrastive training on 400 million image-text pairs. We use the ViT-L/14 variant (\texttt{openai/clip-vit-large-patch14}). These embeddings capture high-level semantic content and have demonstrated strong performance on zero-shot visual recognition tasks. DINOv3 \cite{simeoni2025dinov3} from Meta produces self-supervised visual features trained on curated image data without language supervision. We use the ViT-L variant (\texttt{facebook/dinov3-vitl16-pretrain-lvd1689m}). 

We additionally evaluate concatenated CLIP+DINOv3 embeddings to assess whether combining vision-language and self-supervised representations yields additional discriminative power for distinguishing realistic from unrealistic environmental augmentations.

\subsubsection{Reference Distribution Construction}

For each target condition (fog, rain, snow, night), we construct a reference distribution modelling the expected feature-space characteristics of authentic adverse-condition imagery. Using the real adverse-condition images from the ACDC dataset (excluding 100 held-out images per condition used to establish baseline distances), we fit multivariate Gaussian distributions characterised by condition-specific means $\boldsymbol{\mu}_k$ and covariance matrices $\boldsymbol{\Sigma}_k$ for each condition $k$.

These distributions are computed separately in each embedding space (CLIP, DINOv3, and concatenated), yielding three sets of reference distributions against which augmented images are compared.

\subsubsection{Relative Mahalanobis Distance Metric}

We quantify augmentation quality using relative Mahalanobis distance~\cite{ren2021simple}, an extension of the Mahalanobis distance-based OOD detection framework~\cite{lee2018simple} specifically designed for near-OOD scenarios. Standard Mahalanobis distance measures proximity to class-conditional distributions but can fail when test samples share background features with training data---a common situation when augmented images retain semantics from the original image, while modifying environmental appearance. The relative formulation addresses this limitation by subtracting a background distance term, effectively cancelling out shared features and emphasising condition-discriminative dimensions.

For an augmented image with embedding $\mathbf{x}$ targeting environmental condition $k$, we first compute the Mahalanobis distance to the target distribution:

\begin{equation}
    d_k(\mathbf{x}) = \sqrt{(\mathbf{x} - \boldsymbol{\mu}_k)^\top \boldsymbol{\Sigma}_k^{-1} (\mathbf{x} - \boldsymbol{\mu}_k)}
\end{equation}

We then compute the relative Mahalanobis distance by subtracting the distance to a class-agnostic background distribution fitted to all adverse-condition images:

\begin{equation}
    d_{\text{rel}} = d_k(\mathbf{x}) - d_0(\mathbf{x})
\end{equation}

This relative formulation is particularly well-suited to our task. Synthetic environmental augmentations represent a near-OOD scenario: augmented images share substantial visual semantics with real adverse-condition images (same road scenes, similar object layouts) but differ only in the source of the adverse environmental condition. Standard Mahalanobis distance can be dominated by these shared background features, obscuring meaningful differences in augmentation realism. The relative formulation suppresses contributions from dimensions where augmented and real images are similar, amplifying signal from the adverse condition.

To align interpretive direction with the VLM acceptance metric (where higher values indicate better performance), we report the negated relative distance, $-d_{\mathrm{rel}}$, in all plots and summary statistics. Under this convention, higher values (i.e., values closer to $0$ and less negative) indicate greater proximity to the target condition distribution relative to the background distribution, whereas more negative values indicate greater distance and thus poorer distributional similarity.

We do not normalise $-d_{\mathrm{rel}}$ to a $[0,1]$ range because the relative Mahalanobis score is unbounded and its magnitude depends on the embedding dimensionality and the fitted covariance structure, making any fixed rescaling inherently arbitrary. Since our primary use of this metric is comparative (ranking methods under a fixed embedding model and reference distribution), such normalisation is unnecessary and could obscure the meaning of effect sizes by introducing an additional, ad hoc transformation.

\subsubsection{Embedding Baseline}

We establish expected distances for realistic adverse-condition images using a held-out set of 400 real images (100 per condition). These images are excluded from distribution fitting and serve as a reference for the distances that genuine adverse-condition imagery achieves.

This baseline quantifies the intrinsic variability within each condition category and provides context for interpreting distances computed on augmented images. Notably, because real adverse-condition images are by definition within-distribution, they achieve near-zero distances, making this an upper bound on the baseline. Further analysis is provided in Appendix~\ref{app:embedding_analysis}.

\subsection{Statistical Analysis}

All reported metrics include 95\% bootstrap confidence intervals computed over the sample. Sample sizes comprise 40 images per augmentation method and target condition combination, yielding 160 observations per method across all four conditions. For VLM jury evaluations, this results in 480 total judgments per method (40 images $\times$ 4 conditions $\times$ 3 judges).

We compute Cohen's $\kappa$ to quantify inter-judge agreement for the VLM jury, providing insight into the consistency of perceptual assessments across different vision-language models. Full agreement statistics, including analysis of augmented versus baseline real adverse-condition evaluations, are provided in Appendix~\ref{app:vlm_reliability}.

\subsection{Experimental Controls}

Several design choices ensure rigorous comparison across methods. All augmentation methods transform the same 40 clear-day images, eliminating confounds from source image variability. Generative AI methods receive semantically equivalent prompts describing target conditions.

We examine whether VLM judges exhibit bias when evaluating augmentations produced by models from their own organisation---for example, whether the Gemini judge rates Gemini-generated augmentations more or less favourably than other judges. Results of this cross-evaluation analysis are reported in Appendix~\ref{app:vlm_reliability}.

This methodology enables direct comparison of augmentation paradigms across multiple target conditions while providing baselines to contextualise both the different measures of realism.

\section{Results}

\begin{figure}[htbp]
    \centering
    \includegraphics[width=\linewidth]{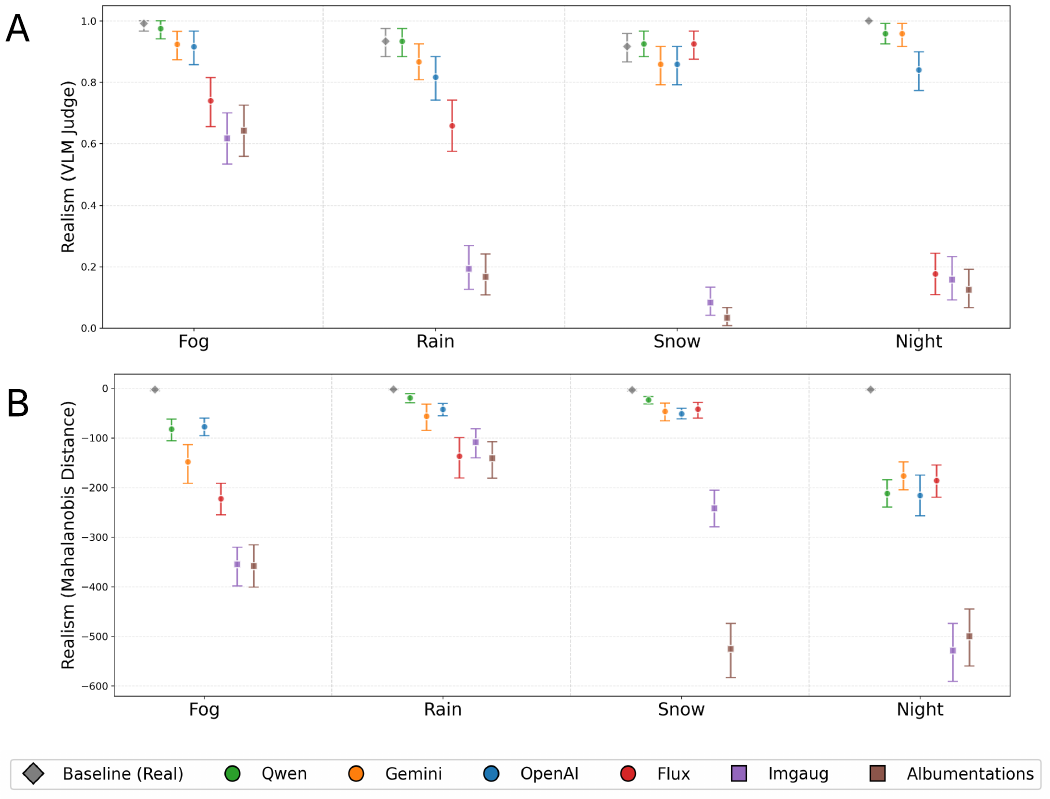}
    \caption{Realism scores by augmentation method and target environmental condition. (A) VLM jury acceptance rates. (B) Negative relative Mahalanobis distance in CLIP embedding space. Grey diamonds indicate baselines from real adverse-condition images. Error bars show 95\% bootstrap confidence intervals. Higher values indicate greater realism.}
    \label{fig:results}
\end{figure}

We present results from both evaluation methods: VLM jury assessments of perceptual realism and embedding-based distributional analysis using relative Mahalanobis distance. Both metrics demonstrate that generative AI image editing models substantially outperform rule-based augmentation methods, with clear patterns across target conditions, but with some notable exceptions. Figure~\ref{fig:results} summarises performance across both evaluation methods.

\subsection{Generative AI Outperforms Rule-based Methods}

The performance gap between paradigms is substantial: The VLM jury shows the best generative method (Qwen, 0.948 [0.927, 0.967]) achieves approximately 3.6 times the acceptance rate of the best rule-based method (imgaug, 0.263 [0.225, 0.299]). All four generative AI methods exceed 0.62 acceptance, while both rule-based libraries remain below 0.27.

Embedding-based analysis corroborates this ranking quantitatively. In CLIP space, the worst-performing generative method (Flux; mean relative Mahalanobis distance $d_{\mathrm{rel}}=99.9$) is still $201.5/99.9 \approx 2.0\times$ closer to the real adverse-condition distributions than the best rule-based method (imgaug; $d_{\mathrm{rel}}=201.5$). The best generative method in CLIP space (OpenAI; $d_{\mathrm{rel}}=46.5$) is $201.5/46.5 \approx 4.3\times$ closer than imgaug. This mirrors the $\approx 0.948/0.263 \approx 3.6\times$ performance gap observed in VLM acceptance rates.
Using CLIP embeddings, the average relative Mahalanobis distance for generative methods ranges from 46.5 (OpenAI) to 99.9 (Flux), compared to 201.5 (imgaug) and 299.6 (albumentations) for Rule-based Methods. DINOv3 and concatenated CLIP+DINOv3 embeddings yield consistent rankings. Full embedding results appear in Appendix~\ref{app:embedding_analysis}.

The Rule-based Methods fail most dramatically on nighttime and snow conditions. For night augmentation, imgaug achieves only 0.158 acceptance and albumentations 0.125, compared to 0.958 for both Qwen and Gemini. Snow augmentation shows similar patterns: albumentations achieves 0.033, essentially complete failure. These conditions require global scene transformations---lighting changes, surface accumulation patterns---that rule-based methods cannot achieve.

\subsection{Exceptions and Notable Patterns}

Two important exceptions qualify the general pattern. First, Rule-based Methods perform respectably on fog augmentation: imgaug achieves 0.617 and albumentations 0.642, approaching the lower-performing generative method (Flux, 0.739). Fog augmentation primarily requires contrast reduction and atmospheric haze---transformations achievable through relatively simple image processing. This represents the only condition where Rule-based Methods approach generative performance. Second, Flux Kontext exhibits strong performance on snow augmentation (0.925, equivalent to the other generative AI methods), however, it is generally much worse than other AI methods. Its nighttime augmentations achieve only 0.176---comparable to Rule-based Methods.

\subsection{Variation Among Generative Methods}

Within the generative paradigm, Qwen and Gemini consistently outperform Flux and OpenAI's GPT-Image-1. Averaged across conditions and judges: Qwen (0.948), Gemini (0.902), OpenAI (0.858), and Flux (0.626). Qwen and Gemini maintain consistently high performance across all conditions, with no acceptance rate below 0.858 for any condition type. By contrast, OpenAI and Flux show greater variability.

The embedding analysis partially corroborates these VLM findings but reveals an interesting discrepancy. In CLIP space, OpenAI augmentations achieve the lowest average distance (46.5), marginally outperforming Qwen (53.7) and Gemini (57.8). This suggests OpenAI augmentations may occupy regions of semantic space closer to real adverse-condition imagery even when VLM judges rate them as less perceptually realistic. We discuss potential explanations for this divergence in Section~\ref{sec:discussion}.

\subsection{Judge Consistency and Cross-Company Evaluation}

The three VLM judges exhibit substantial variation in acceptance rates. Claude (0.759) and GPT-4o (0.722) produce similar overall scores, while Gemini 2.5 Pro is considerably more conservative (0.438). Inter-judge agreement is moderate: Cohen's $\kappa$ between Claude and GPT-4o reaches 0.69, but agreement with Gemini is lower ($\kappa = 0.37$ for Claude--Gemini, $\kappa = 0.45$ for GPT--Gemini). Detailed inter-judge analysis appears in Appendix~\ref{app:vlm_reliability}.

A natural concern is whether judges exhibit bias toward augmentations produced by models from their own organisation. We find no evidence of positive bias; if anything, the pattern suggests self-criticism. When the Gemini judge evaluates Gemini-generated augmentations, acceptance rates are \textit{lower} than the cross-judge average: rain drops from 0.867 overall to 0.650 under Gemini evaluation, and snow from 0.858 to 0.650. GPT-4o evaluating OpenAI augmentations produces scores similar to the overall average, showing no discernible favouritism.

To test robustness, we examined rankings excluding same-company judge--method pairs. The method rankings remain unchanged: Qwen and Gemini continue to outperform OpenAI and Flux regardless of which judge combinations are used. We additionally assessed whether Gemini's conservative scoring drives our conclusions by computing rankings under alternative jury compositions (Gemini-only, Claude+GPT-4o only, and full jury). Method rankings remained stable across all configurations, with the top four methods maintaining identical ordering; the primary effect was on absolute acceptance levels rather than relative ranking (full analysis in Appendix~\ref{app:vlm_reliability}). This provides confidence that the observed performance differences reflect genuine differences rather than judge bias.

\subsection{Comparison to Ground Truth Baselines}

To contextualise synthetic augmentation scores, we established baselines using genuine adverse-condition imagery from the ACDC dataset. The VLM jury evaluated 40 real images per condition using identical criteria.

Real adverse-condition images do not achieve perfect acceptance. Night imagery reaches 1.000, and fog 0.992, but rain achieves only 0.933 and snow 0.917. These imperfect baselines have important implications for interpreting synthetic augmentation performance. For rain, Qwen matches the baseline exactly (0.933). For snow, Flux and Qwen (both 0.925) actually exceed the real baseline (0.917). For fog, Qwen (0.975) falls just short of baseline (0.992), likely within noise given overlapping confidence intervals. Only nighttime augmentations fall meaningfully short: the best methods (Gemini, Qwen at 0.958) approach but do not reach the perfect 1.000 baseline.

The embedding baseline tells a different story. Held-out real adverse-condition images achieve near-zero relative Mahalanobis distances to their own condition distributions (0.2--3.2 for CLIP, 2.0--3.3 for DINOv3), as expected as these are in fact within-distribution samples, taken from the same ACDC dataset. Even the best augmentation methods remain substantially further from these distributions (6.3--185.0 for CLIP). This gap is inherent to the evaluation design: real images are by definition within-distribution, while augmented images transform clear-day scenes that differ structurally from genuine adverse-condition imagery. The embedding baseline thus establishes an upper bound.

We did not manually verify every baseline image beyond using the ACDC-provided condition labels; baseline images were randomly sampled from the dataset by their labelled condition. Consequently, the fact that real adverse-condition imagery does not reach 100\% acceptance (e.g., 0.933 for rain and 0.917 for snow) should not be interpreted as judge ``error'' alone. Several non-exclusive factors could contribute: (i) genuine ambiguity and within-class variability in real-world conditions (e.g., light rain or early/patchy snowfall that does not produce unambiguous visual cues), (ii) borderline frames where the labelled condition is present but visually subtle (e.g., shortly after precipitation), and (iii) potential label noise in the underlying dataset annotations. We therefore treat these baseline acceptance rates as practical ceilings for our automated protocol rather than as ground-truth upper bounds.

\subsection{Failure Mode Analysis}

To understand the nature of augmentation failures, we classified VLM judge rejection reasons into semantic preservation failures and realism failures using a secondary LLM jury (see Appendix~\ref{app:failure_analysis} for methodology).

Realism failures dominate overall: 80.0\% of rejections cite unrealistic appearance alone, while only 8.6\% cite semantic non-preservation alone. However, failure modes differ markedly between paradigms. Rule-based Methods fail almost exclusively due to unrealistic appearance (97.5\% realism-only for imgaug) while perfectly preserving semantic content. Top-performing generative methods show the inverse pattern: OpenAI and Qwen failures are predominantly semantic (74.0\% and 66.7\% semantic-only), indicating these methods produce realistic-looking effects but sometimes alter scene content. This reveals a fundamental trade-off between semantic fidelity and visual realism.

\subsection{Can Absolute Scores Be Trusted?}

Can VLM jury scores be interpreted as absolute measures of augmentation quality? This claim must be caveated in the following ways.

First, inter-judge agreement is only moderate ($\kappa = 0.37$--0.69), indicating judges apply different standards. The Gemini judge rejects augmentations that Claude and GPT-4o accept, yet achieves the highest acceptance rate on real adverse-condition images (0.994 versus 0.919--0.969 for other judges). This suggests Gemini applies stricter criteria specifically to synthetic imagery, perhaps due to a genuine lack of realism, not picked up by the other judges.

Second, we must consider the possibility that synthetic augmentations depict more extreme adverse conditions than naturally occurring imagery, which could inflate their acceptance scores. Real adverse-condition baselines are imperfect---genuine rain images achieve only 0.933 acceptance, and snow 0.917---likely reflecting the natural variability in condition severity. Light rainfall or early-stage snow accumulation may not satisfy judges' expectations for unambiguous condition presence. If synthetic methods tend to produce more pronounced effects, direct comparison to real baselines requires caution. A method achieving 0.90 on rain may be performing at approximately 96\% of the empirical ceiling, or it may be exceeding real baselines precisely because it generates more pronounced conditions than the mild adverse weather present in some genuine imagery.

We feel most comfortable interpreting VLM jury scores as \textit{relative} measures suitable for ranking methods within a consistent evaluation framework, rather than an \textit{absolute} quality measure. The consistent ranking across judges, conditions, and evaluation methods supports confidence in comparative conclusions: generative AI methods produce more realistic augmentations than rule-based approaches, and Qwen and Gemini outperform Flux and OpenAI within the generative paradigm.

\subsection{Condition-Specific Patterns}

Performance varies substantially across target conditions, with patterns largely consistent between evaluation methods.

\textbf{Fog} emerges as the easiest condition to simulate. All four generative methods exceed 0.73 VLM acceptance, and even Rule-based Methods achieve their best performance here.

\textbf{Snow and rain} present moderate difficulty. Generative methods achieve 0.82--0.93 acceptance for both conditions, while Rule-based Methods largely fail (0.03--0.19). These conditions require plausible precipitation rendering and surface appearance changes that demand more fine-grained manipulation.

\textbf{Night} Nighttime transformation results are particularly encouraging for top generative methods, with Qwen and Gemini achieving 0.958 VLM acceptance despite the global illumination changes required. Flux's poor performance (0.176) appears attributable to excessive darkening that obscures scene semantics rather than fundamental limitations of generative approaches. The striking divergence between high VLM acceptance and large embedding distances for nighttime augmentations suggests a limitation of the embedding-based evaluation. CLIP and DINOv3 representations may be dominated by global colour and luminance statistics that distinguish day from night, failing to capture the contextual appropriateness of lighting that VLM judges recognise---headlights, streetlamps, and darkness in expected regions. This indicates that embedding space analysis may not reliably assess synthetic nighttime augmentations, even when perceptual quality is high. We discuss the divergence in more detail in Appendix~\ref{app:embedding_analysis}.

Notably, embedding-based rankings diverge from VLM rankings for some method--condition combinations. OpenAI augmentations achieve lower CLIP distances than Qwen despite lower VLM acceptance. Flux nighttime augmentations show moderate embedding distances (219.8 CLIP) despite very low VLM acceptance (0.176). These discrepancies suggest that proximity in semantic embedding space does not perfectly predict perceptual realism judgments---a point we explore further in Section~\ref{sec:discussion}.

\section{Discussion}
\label{sec:discussion}

The key practical implication is that rule-based libraries are unsuitable when realism is required for safety-critical evaluation. The failure mode analysis reveals a fundamental trade-off between augmentation paradigms. Rule-based Methods preserve semantic content perfectly but produce obviously artificial effects---when they fail, it is almost exclusively due to unrealistic appearance (97.5\% for imgaug). Generative methods achieve visual realism but at the cost of semantic preservation---OpenAI and Qwen failures are predominantly due to altered scene content (74.0\% and 66.7\% semantic-only). This suggests practitioners must choose between methods that never alter content but look artificial, versus methods that appear realistic but may inadvertently modify scene semantics. For safety-critical evaluation, where both properties matter, the top generative methods (Qwen, Gemini) offer the best balance, achieving high acceptance rates while maintaining lower failure counts overall.

\subsection{Divergence Between Evaluation Modalities}

The partial divergence between VLM jury scores and embedding-based distances warrants attention. OpenAI augmentations achieve the lowest CLIP distances despite ranking third in perceptual acceptance. This suggests the two metrics capture different aspects of realism: embedding proximity measures statistical similarity to real adverse-condition distributions, while VLM judges assess perceptual convincingness. An augmentation may occupy the correct region of semantic space statistically yet contain artifacts that human-like perception recognises as unrealistic---or conversely, appear perceptually convincing while being statistically distant from the reference distribution. To investigate which metric better captures meaningful realism when they disagree, we conducted targeted qualitative analysis of cases with maximal metric divergence; results in Appendix~\ref{app:metric_divergence} suggest VLM judgments align more reliably with intuitive assessments of augmentation quality.

The divergence is most pronounced for nighttime augmentation. All methods achieve embedding distances 65--110$\times$ greater than baseline, yet VLM judges rate Qwen and Gemini outputs as close to genuine adverse condition imagery (0.958 acceptance versus 1.000 baseline). This striking divergence suggests a limitation of the embedding-based evaluation rather than poor augmentation quality.

CLIP and DINOv3 embeddings yield consistent method rankings but differ in absolute distances. CLIP's vision-language training produces tighter condition clusters, while DINOv3's self-supervised features show greater sensitivity to low-level texture differences. Neither embedding space is inherently superior; they capture complementary aspects of visual similarity. Full comparison of embedding spaces is provided in Appendix~\ref{app:embedding_analysis}.

\subsection{Baseline Reliability}

The two baselines serve different interpretive functions. The VLM jury baseline reveals that real adverse-condition images achieve imperfect acceptance---only 0.917 for snow and 0.933 for rain. This likely reflects genuine ambiguity in condition appearance: light snowfall may be indistinguishable from clear conditions, and recently-dried roads may not exhibit obvious rain signatures. The imperfect baseline establishes practical ceilings against which synthetic methods should be judged. By this standard, Qwen's rain augmentation (0.933) matches real imagery, and both Qwen and Flux exceed the snow baseline (0.925 versus 0.917).

The embedding baseline is less informative by design. Held-out real adverse-condition images achieve near-zero relative Mahalanobis distances because they are, by definition, within-distribution samples. Augmented images transform clear-day scenes whose underlying structure differs from genuine adverse-condition imagery, creating an inherent gap that no augmentation method can close entirely. The embedding baseline thus represents an upper-bound ceiling rather than a practical target.

\subsection{VLM Jury Reliability}

The moderate inter-judge agreement ($\kappa = 0.37$--0.69) raises questions about VLM jury reliability. Gemini's pattern is particularly striking: it achieves the highest acceptance on real images (0.994) yet the lowest on augmented images (0.438), suggesting heightened sensitivity to synthetic artifacts. The absence of cross-company bias---and Gemini's self-critical tendency---provides some confidence that scores reflect genuine quality differences rather than systematic favouritism.

We recommend interpreting VLM scores as relative rankings rather than absolute quality measures at this point. The consistent ordering across judges and conditions supports comparative conclusions even where absolute calibration remains uncertain.

\subsection{On Automated Evaluation Without Human Annotators}

A potential limitation of this study is our reliance on automated evaluation without human annotator validation. However, we argue this reflects a principled trade-off rather than mere convenience: comprehensive evaluation across multiple augmentation methods, conditions, and datasets quickly requires thousands of judgments, which is difficult to sustain with human annotators at the iteration speed required for method development and monitoring. By contrast, model-based judging enables scalable, repeatable evaluation under a fixed protocol, and prior work shows that strong LLM-based judges can align with human preference signals at levels comparable to human--human agreement in analogous settings \cite{zheng2023judging}. Related work in image editing evaluation similarly reports strong correlations between multimodal model scores and human ratings across several editing tasks \cite{huang2025diffusion}.

Human judgments are also not an infallible gold standard. Even among experts, inter-rater agreement can be only fair-to-moderate, reflecting subjectivity and ambiguous edge cases, and annotation can be influenced by cognitive and procedural effects (e.g., anchoring) \cite{plenitz2023inconsistent,obuchowicz2020mriquality,parmar2024blindspots}. Rather than treating any single evaluator as an oracle, we therefore seek corroboration through methodological diversity: (i) a multi-model VLM jury (GPT-4o, Claude, Gemini), and (ii) an independent embedding-based analysis using non-generative representation models (CLIP and DINOv3). The central conclusion---that generative methods substantially outperform rule-based methods---is supported by convergence between these independent evaluation modalities. This mirrors the broader wisdom-of-crowds rationale for aggregating diverse perspectives \cite{surowiecki2004wisdom} and is consistent with evidence that aggregated model judgments can rival human crowd performance \cite{schoenegger2024siliconcrowd}. We present this framework primarily for relative comparison; future work should validate it directly against human studies.

\subsection{Limitations and Generalisability}

Several limitations constrain generalisability. Our evaluation uses a single dataset (ACDC) with specific camera characteristics and geographic contexts; augmentation performance may differ on imagery from other sensors or regions. The 40-image test set, while sufficient for ranking methods, limits statistical power for detecting smaller effect sizes.

Finally, our evaluation relies on automated metrics. While prior work shows that multimodal model judgments can correlate with human perceptual evaluations in related domains, validation against human annotators would strengthen confidence in our findings in this specific setting. The scalability advantages of automated evaluation must be weighed against general discomfort with forgoing human validation.

\section{Conclusion}

We presented a systematic evaluation framework for assessing environmental augmentation realism, comparing rule-based and generative AI methods across four conditions using complementary automated metrics. Our findings provide clear practical guidance for practitioners building evaluation datasets for safety-critical applications.

Generative AI image editing models substantially outperform traditional rule-based augmentation libraries. The performance gap is approximately fourfold: Qwen achieves 0.948 VLM acceptance compared to 0.263 for imgaug. This disparity is consistent across both evaluation modalities and most conditions, indicating that rule-based augmentation libraries are unsuitable for safety-critical evaluation where realism matters. Among generative methods, Qwen and Gemini emerge as consistently strong performers, maintaining high acceptance rates across all conditions.

Target conditions differ substantially in augmentation difficulty. Fog proves amenable even to simple image processing techniques, while snow and rain require the scene understanding that generative models provide. Nighttime transformation yields encouraging results for the top generative methods, with near-ceiling VLM acceptance; however, embedding analysis reveals all methods produce outputs far from genuine night imagery distributions, likely reflecting a limitation of embedding-based evaluation for nighttime scenes rather than poor augmentation quality.

The partial divergence between VLM jury scores and embedding distances indicates these metrics capture complementary aspects of realism. We recommend using both evaluation modalities and interpreting scores as relative rankings rather than absolute quality certifications. Qualitative analysis of cases where metrics disagree suggests VLM judgments may better capture perceptual realism, though embedding distances remain valuable for detecting distributional shifts. The imperfect baseline acceptance for real adverse-condition imagery (0.917--1.000) establishes practical ceilings that contextualise synthetic augmentation performance.

Failure mode analysis reveals that rule-based and generative methods fail for opposite reasons---unrealistic appearance versus semantic alteration, respectively---suggesting a fundamental trade-off that practitioners should consider when selecting augmentation approaches.

Future work should extend evaluation to additional datasets and domains, assess control alongside realism, and validate automated metrics against human annotators. As AI systems proliferate across safety-critical applications, scalable evaluation methodologies like those presented here will become increasingly essential for ensuring adequate pre-deployment testing coverage.

\clearpage
\bibliographystyle{plainnat}
\bibliography{references}

\appendix
\clearpage
\section{VLM Jury Analysis}
\label{app:vlm_reliability}

This appendix provides detailed analysis of the VLM jury evaluation methodology, including judge configuration, inter-judge agreement, baseline calibration, and investigation of potential cross-company biases.

\subsection{Judge Configuration}

\subsubsection{Prompt Templates}

All VLM judges received standardised prompts to ensure consistent evaluation criteria. For augmented image evaluation, judges received image pairs (original clear-day image and augmented version) with the following task specification:

\begin{quote}
\textit{``Evaluate whether the [condition] augmentation of an image looks realistic and preserves the semantics from the original (clear day) image.''}
\end{quote}

Judges assessed two criteria that both had to be satisfied for a positive decision:

\begin{enumerate}
    \item \textbf{Semantic Preservation}: Other than the target condition being added, is the image otherwise the same? Are the objects and semantics preserved, with the exception of elements naturally obscured by the target condition?
    \item \textbf{Condition Realism}: Is the target condition that has been simulated realistic? Does the image actually look like it was taken in that condition?
\end{enumerate}

Condition-specific guidance was provided for each target. Snow augmentations should exhibit evidence of falling or accumulated snow. Rain augmentations should show precipitation or wet surfaces. Fog augmentations should demonstrate atmospheric obscuration with reduced visibility. Night transformations should feature appropriate darkness and artificial lighting characteristics.

For baseline evaluation of real adverse-condition images, judges received single images with the task: \textit{``Evaluate whether this image shows realistic [condition] conditions.''} This focused solely on condition realism without semantic preservation assessment.

All prompts required step-by-step reasoning followed by a JSON output containing an ``explanation'' field and a binary ``decision'' field (true/false).

\subsubsection{Model-Specific Hyperparameters}

Table~\ref{tab:judge_config} summarises the configuration for each VLM judge.

\begin{table}[htbp]
\centering
\caption{VLM judge model configurations.}
\label{tab:judge_config}
\begin{tabular}{llccc}
\hline
\textbf{Judge} & \textbf{Model} & \textbf{Max Tokens} & \textbf{Temperature} & \textbf{Reasoning Mode} \\
\hline
GPT-4o & gpt-4o & 2048 & Default & None \\
Claude & claude-sonnet-4-20250514 & 2048 & Default & Extended (1024 tokens) \\
Gemini & gemini-2.5-pro & 2048 & Default & Dynamic \\
\hline
\end{tabular}
\end{table}

Claude employed extended thinking with a 1024-token budget. Gemini used dynamic thinking, automatically adjusting reasoning depth as needed. GPT-4o operated without explicit reasoning mode. Images were compressed to under 3.5 MB for Claude's input constraints.

\subsection{Inter-Judge Agreement}

We computed Cohen's $\kappa$ to quantify agreement between judge pairs. Table~\ref{tab:agreement} presents agreement statistics for both augmented and real adverse-condition image evaluations.

\begin{table}[htbp]
\centering
\caption{Inter-judge agreement (Cohen's $\kappa$) for augmented and real adverse-condition images.}
\label{tab:agreement}
\begin{tabular}{lcc}
\hline
\textbf{Judge Pair} & \textbf{Augmented Images} & \textbf{Real Images} \\
\hline
Claude--GPT-4o & 0.69 & 0.19 \\
Claude--Gemini & 0.37 & 0.33 \\
GPT-4o--Gemini & 0.45 & 0.13 \\
\hline
\end{tabular}
\end{table}

Claude and GPT-4o exhibited substantial agreement ($\kappa = 0.69$) on augmented images, suggesting similar evaluation standards. Agreement with Gemini was notably lower ($\kappa = 0.37$--$0.45$), indicating systematically different criteria. Interestingly, agreement on real adverse-condition images was lower across all pairs, suggesting judges apply different thresholds for ``perfect'' realism even on genuine imagery.

\subsection{Judge Performance Characteristics}

Table~\ref{tab:judge_performance} presents overall acceptance rates by judge across all augmentation methods and target conditions.

\begin{table}[htbp]
\centering
\caption{Overall judge acceptance rates with 95\% bootstrap confidence intervals.}
\label{tab:judge_performance}
\begin{tabular}{lccc}
\hline
\textbf{Judge} & \textbf{Augmented Images} & \textbf{Real Images} & \textbf{N} \\
\hline
Claude & 0.759 [0.732, 0.787] & 0.969 [0.944, 0.994] & 956/160 \\
GPT-4o & 0.722 [0.692, 0.751] & 0.919 [0.875, 0.956] & 957/160 \\
Gemini & 0.438 [0.407, 0.471] & 0.994 [0.981, 1.000] & 960/160 \\
\hline
\end{tabular}
\end{table}

A notable pattern emerges: Gemini scores augmented images substantially lower than other judges (0.438 vs.\ 0.722--0.759) yet achieves the highest acceptance rate on real adverse-condition images (0.994). This suggests Gemini applies particularly stringent criteria to synthetic augmentations while accurately recognising genuine adverse-condition imagery. The discrepancy implies heightened sensitivity to subtle artifacts or imperfections introduced by augmentation methods.

\subsection{Baseline Calibration}

Table~\ref{tab:baseline} presents baseline acceptance rates for real adverse-condition images by condition, establishing empirical ceilings for synthetic augmentation performance.

\begin{table}[htbp]
\centering
\caption{Baseline acceptance rates for real adverse-condition images by condition (N=120 per condition).}
\label{tab:baseline}
\begin{tabular}{lc}
\hline
\textbf{Condition} & \textbf{Acceptance Rate [95\% CI]} \\
\hline
Night & 1.000 [1.000, 1.000] \\
Fog & 0.992 [0.975, 1.000] \\
Rain & 0.933 [0.883, 0.975] \\
Snow & 0.917 [0.858, 0.958] \\
\hline
\end{tabular}
\end{table}

Real adverse-condition images do not achieve universal acceptance. Rain (0.933) and snow (0.917) baselines fall notably short of perfection, indicating that some genuine adverse-condition images fail to convince VLM judges. This has important implications: a synthetic method achieving 0.93 acceptance on rain augmentation performs at the empirical ceiling established by real imagery. The imperfect baselines suggest either conservative judge standards or inherent ambiguity in condition appearance even for authentic images.

\subsection{Cross-Company Bias Analysis}

A natural concern is whether VLM judges exhibit preferential treatment toward augmentations produced by models from their parent organisation. We investigated this by comparing acceptance rates when judges evaluate augmentations from their own company versus the cross-judge average.

\subsubsection{OpenAI Judge Evaluating OpenAI Augmentations}

Table~\ref{tab:openai_bias} compares GPT-4o's evaluation of OpenAI-generated augmentations against overall scores.

\begin{table}[htbp]
\centering
\caption{OpenAI augmentation scores: GPT-4o judge vs.\ overall average.}
\label{tab:openai_bias}
\begin{tabular}{lcc}
\hline
\textbf{Condition} & \textbf{GPT-4o Score} & \textbf{Overall Score} \\
\hline
Fog & 0.925 & 0.916 \\
Night & 0.850 & 0.840 \\
Rain & 0.825 & 0.817 \\
Snow & 0.850 & 0.858 \\
\hline
\end{tabular}
\end{table}

GPT-4o's scores for OpenAI augmentations closely match the cross-judge averages across all conditions. No evidence of positive bias is apparent; differences fall within expected sampling variation.

\subsubsection{Gemini Judge Evaluating Gemini Augmentations}

Table~\ref{tab:gemini_bias} presents a contrasting pattern for Gemini's self-evaluation.

\begin{table}[htbp]
\centering
\caption{Gemini augmentation scores: Gemini judge vs.\ overall average.}
\label{tab:gemini_bias}
\begin{tabular}{lcc}
\hline
\textbf{Condition} & \textbf{Gemini Score [95\% CI]} & \textbf{Overall Score} \\
\hline
Fog & 0.800 [0.675, 0.925] & 0.924 \\
Night & 0.875 [0.750, 0.975] & 0.958 \\
Rain & 0.650 [0.500, 0.800] & 0.867 \\
Snow & 0.650 [0.500, 0.775] & 0.858 \\
\hline
\end{tabular}
\end{table}

Gemini exhibits self-critical behaviour: it scores its own augmentations \textit{lower} than the cross-judge average across all conditions. The discrepancy is substantial for rain (0.650 vs.\ 0.867) and snow (0.650 vs.\ 0.858). This pattern suggests no positive bias toward same-company models; if anything, Gemini applies stricter standards to its own outputs.

\subsubsection{Robustness Check}

To verify that cross-company dynamics do not distort method rankings, we recomputed rankings excluding same-company judge--method pairs. The ordering remained unchanged: Qwen and Gemini continued to outperform OpenAI and Flux across all evaluation configurations. This provides confidence that observed performance differences reflect genuine quality variation rather than systematic judge bias.

\subsection{Jury Composition Robustness}

Given Gemini's substantially lower acceptance rates on augmented images (0.438 vs.\ 0.722--0.759 for other judges), we assessed whether this conservative judge drives our comparative conclusions by computing method rankings under alternative jury compositions.

Table~\ref{tab:jury_composition} presents acceptance rates under three configurations: Gemini-only, non-Gemini (Claude and GPT-4o only), and the full three-judge jury.

\begin{table}[htbp]
\centering
\caption{Method acceptance rates by jury composition. Rankings shown in parentheses.}
\label{tab:jury_composition}
\begin{tabular}{lccc}
\hline
\textbf{Method} & \textbf{Gemini-only} & \textbf{Non-Gemini} & \textbf{Full Jury} \\
\hline
Qwen & 0.850 (1) & 0.997 (1) & 0.948 (1) \\
Gemini & 0.744 (2) & 0.981 (2) & 0.902 (2) \\
OpenAI & 0.625 (3) & 0.975 (3) & 0.858 (3) \\
Flux & 0.381 (4) & 0.748 (4) & 0.626 (4) \\
imgaug & 0.013 (5) & 0.389 (5) & 0.263 (5) \\
albumentations & 0.013 (5) & 0.356 (6) & 0.242 (6) \\
\hline
\end{tabular}
\end{table}

The top four methods (Qwen, Gemini, OpenAI, Flux) maintain identical rank ordering across all three jury configurations. Rule-based methods (imgaug, albumentations) remain in the bottom tier regardless of composition, though their relative ordering is effectively tied under Gemini-only evaluation (both at 1.3\% acceptance).

Gemini's conservatism uniformly lowers all acceptance rates---Qwen drops from 99.7\% under non-Gemini evaluation to 85.0\% under Gemini-only---but preserves the comparative ranking structure. This stability provides confidence that our conclusions about method performance are robust to individual judge characteristics. The primary effect of jury composition is on absolute acceptance levels, not relative ordering.

\subsection{Summary}

The VLM jury demonstrates moderate inter-judge agreement, with Claude and GPT-4o showing substantial concordance while Gemini applies systematically stricter criteria. Despite this variation, method rankings remain stable across jury compositions, indicating that comparative conclusions are robust to individual judge idiosyncrasies. Baseline calibration reveals that real adverse-condition images achieve imperfect acceptance (0.917--1.000), establishing practical ceilings for synthetic augmentation evaluation. Cross-company bias analysis finds no evidence of favouritism; Gemini's self-critical pattern suggests judges may apply heightened scrutiny to familiar outputs. We recommend interpreting VLM jury scores as relative measures for ranking methods within the evaluation framework rather than absolute quality certifications.

\clearpage
\section{Embedding-based Distributional Analysis}
\label{app:embedding_analysis}

This appendix provides detailed analysis of the embedding-based evaluation methodology, comparing CLIP and DINOv3 representations, examining relationships to baseline distributions, and identifying conditions where augmentation methods systematically struggle.

\subsection{CLIP versus DINOv3 Embeddings}

The two embedding models capture different aspects of visual content due to their distinct training objectives. CLIP produces 768-dimensional embeddings trained on 400 million image-text pairs, encoding semantic and conceptual relationships aligned with natural language. DINOv3 produces 1024-dimensional embeddings through self-supervised learning on curated image data, capturing visual structure and low-level patterns without language supervision.

Table~\ref{tab:embedding_ranking} presents method rankings by average relative Mahalanobis distance across all target conditions.

\begin{table}[htbp]
\centering
\caption{Method performance by embedding space (average relative Mahalanobis distance across conditions; lower is better).}
\label{tab:embedding_ranking}
\begin{tabular}{lccc}
\hline
\textbf{Method} & \textbf{CLIP} & \textbf{DINOv3} & \textbf{CLIP+DINOv3} \\
\hline
OpenAI & 46.5 & 96.8 & 178.1 \\
Qwen & 53.7 & 85.4 & 188.9 \\
Gemini & 57.8 & 105.9 & 231.8 \\
Flux & 99.9 & 144.5 & 295.2 \\
imgaug & 201.5 & 307.1 & 689.1 \\
albumentations & 299.6 & 379.3 & 886.8 \\
\hline
\end{tabular}
\end{table}

Both embedding spaces yield consistent relative rankings: generative methods substantially outperform rule-based approaches, with Qwen and OpenAI achieving the lowest distances. However, CLIP produces lower absolute distances than DINOv3, suggesting tighter clustering of condition representations in vision-language space. The concatenated CLIP+DINOv3 embeddings amplify separation between methods without altering rankings.

One notable discrepancy emerges: OpenAI achieves the lowest CLIP distance (46.5) despite ranking third in VLM jury acceptance (0.858). This suggests OpenAI augmentations occupy regions of semantic embedding space statistically similar to real adverse-condition imagery, even when perceptual judges rate them as less convincing than Qwen or Gemini outputs.

\subsection{Relationship to Baseline Distributions}

Table~\ref{tab:embedding_baseline} presents baseline distances for held-out real adverse-condition images (100 per condition) to their own condition distributions.

\begin{table}[htbp]
\centering
\caption{Baseline relative Mahalanobis distances for real adverse-condition images [95\% CI].}
\label{tab:embedding_baseline}
\begin{tabular}{lccc}
\hline
\textbf{Condition} & \textbf{CLIP} & \textbf{DINOv3} & \textbf{CLIP+DINOv3} \\
\hline
Fog & 3.2 [-0.2, 8.8] & 2.0 [0.2, 4.2] & 8.2 [4.1, 14.3] \\
Night & 2.0 [0.3, 4.1] & 2.4 [1.1, 3.8] & 9.8 [5.6, 15.8] \\
Rain & 0.2 [-0.6, 1.2] & 2.0 [1.1, 3.0] & 7.2 [4.8, 9.9] \\
Snow & 1.6 [-0.3, 3.9] & 3.3 [1.5, 5.3] & 9.4 [6.2, 13.3] \\
\hline
\end{tabular}
\end{table}

Real adverse-condition images achieve near-zero relative distances, as expected for within-distribution samples. This creates an inherent asymmetry: the baseline represents an unattainable ceiling rather than a practical target, since augmented images transform clear-day scenes that differ structurally from genuine adverse-condition imagery.

Table~\ref{tab:best_aug_vs_baseline} compares the best-performing augmentation method per condition against baseline distances.

\begin{table}[htbp]
\centering
\caption{Best augmentation method distances versus baseline (CLIP embeddings).}
\label{tab:best_aug_vs_baseline}
\begin{tabular}{lcccc}
\hline
\textbf{Condition} & \textbf{Best Method} & \textbf{Distance} & \textbf{Baseline} & \textbf{Ratio} \\
\hline
Snow & Qwen & 6.3 & 1.6 & 3.9$\times$ \\
Rain & Qwen & 9.1 & 0.2 & 45.5$\times$ \\
Fog & Qwen & 14.5 & 3.2 & 4.5$\times$ \\
Night & OpenAI & 130.3 & 2.0 & 65.2$\times$ \\
\hline
\end{tabular}
\end{table}

Even the best augmentation methods remain substantially further from condition distributions than real imagery. The ratio varies considerably: snow and fog augmentations achieve distances approximately 4$\times$ baseline, while rain shows a 45$\times$ gap despite small absolute differences. Night augmentations exhibit the largest gap at 65$\times$ baseline.

\subsection{Nighttime: Systematic Challenge}

Nighttime augmentation proves systematically difficult across all methods and embedding spaces. Table~\ref{tab:night_distances} presents night-specific distances.

\begin{table}[htbp]
\centering
\caption{Nighttime augmentation distances by method and embedding space.}
\label{tab:night_distances}
\begin{tabular}{lccc}
\hline
\textbf{Method} & \textbf{CLIP} & \textbf{DINOv3} & \textbf{CLIP+DINOv3} \\
\hline
OpenAI & 130.3 & 223.8 & 444.5 \\
Gemini & 135.0 & 179.8 & 558.9 \\
Qwen & 185.0 & 216.5 & 596.7 \\
Flux & 219.8 & 188.6 & 499.8 \\
\hline
\textit{Baseline} & \textit{2.0} & \textit{2.4} & \textit{9.8} \\
\hline
\end{tabular}
\end{table}

All generative methods achieve night distances 65--110$\times$ greater than baseline in CLIP space, compared to 4--45$\times$ for other conditions. This gap persists in DINOv3 (75--93$\times$ baseline) and concatenated embeddings (45--61$\times$ baseline).

Interestingly, embedding rankings for night diverge from VLM jury results. OpenAI and Gemini achieve the lowest embedding distances despite Qwen and Gemini achieving the highest VLM acceptance (both 0.958). This suggests embedding models capture lighting and colour distribution differences that VLM judges either overlook or weight differently in perceptual assessment.

\subsection{Summary}

CLIP and DINOv3 embeddings yield consistent method rankings, with CLIP providing better absolute separation. The embedding baseline establishes an unattainable ceiling since real adverse-condition images are by definition within-distribution. Nighttime augmentation emerges as a systematic challenge: all methods achieve distances 65--110$\times$ baseline, substantially larger than the 4--45$\times$ gaps observed for other conditions. The divergence between embedding distances and VLM acceptance for night suggests these evaluation modalities capture different aspects of augmentation quality.

\clearpage
\section{Failure Mode Analysis}
\label{app:failure_analysis}

To understand why augmentations fail, we conducted a secondary analysis classifying VLM judge rejection reasons into distinct failure categories. This analysis provides insight into the fundamental differences between augmentation paradigms and identifies specific weaknesses of each method.

\subsection{Two-Stage Evaluation Framework}

Our failure analysis employs a two-stage approach. In the first stage (described in Section~3.3), VLM judges evaluated augmented images and produced binary decisions with textual explanations for rejections. In the second stage, we classified these rejection reasons using a separate LLM jury.

For each failed augmentation, we analyzed the VLM judge's textual explanation to determine whether the failure was due to: (1) \textit{semantic non-preservation}---objects or scene content changed from the original image, or (2) \textit{realism failure}---the augmentation appears artificial or unconvincing. These categories are treated as independent binary classifications; a failure may exhibit one, both, or neither issue.

\subsection{LLM Judge Configuration}

Three LLM judges independently classified each failure reason:

\begin{table}[htbp]
\centering
\caption{LLM judge configurations for failure classification.}
\label{tab:llm_judge_config}
\begin{tabular}{llc}
\hline
\textbf{Judge} & \textbf{Model} & \textbf{Configuration} \\
\hline
Claude & claude-sonnet-4-20250514 & thinking\_budget=1024 \\
Gemini & gemini-2.5-pro & thinking\_budget=-1 (dynamic) \\
GPT & gpt-4o & default \\
\hline
\end{tabular}
\end{table}

Each judge received the VLM's rejection reason and was asked to determine whether it indicated semantic non-preservation, realism failure, or both. For semantic preservation, judges assessed whether the reason indicated objects being added, removed, or changed; scene layout being altered; or key elements being distorted or replaced. For realism, judges assessed whether the reason indicated artificial-looking weather effects, inconsistent lighting, unnatural textures, or visual artifacts.

This produced six judgments per failure (3 judges $\times$ 2 categories), yielding 3,108 total classifications across 1,036 unique failures.

\subsection{Failure Type Distribution}

Table~\ref{tab:failure_distribution} presents the overall distribution of failure types, aggregated across all LLM judge classifications.

\begin{table}[htbp]
\centering
\caption{Distribution of failure types across all rejected augmentations (N=3,108 classifications).}
\label{tab:failure_distribution}
\begin{tabular}{lrc}
\hline
\textbf{Failure Type} & \textbf{Count} & \textbf{Percentage} \\
\hline
Realism only & 2486 & 80.0\% \\
Both semantic and realism & 311 & 10.0\% \\
Semantic only & 267 & 8.6\% \\
Neither (ambiguous) & 44 & 1.4\% \\
\hline
\textbf{Total} & 3108 & 100.0\% \\
\hline
\end{tabular}
\end{table}

Realism failures dominate: 90.0\% of rejections cite unrealistic appearance (alone or combined with semantic issues), while only 18.6\% cite semantic non-preservation. The small ``neither'' category (1.4\%) represents cases where the VLM judge's textual explanation was ambiguous and did not clearly indicate either failure type.

\subsection{Failure Modes by Augmentation Method}

Table~\ref{tab:failure_by_method} reveals a stark dichotomy between augmentation paradigms.

\begin{table}[htbp]
\centering
\caption{Failure type distribution by augmentation method. Percentages are row-wise.}
\label{tab:failure_by_method}
\begin{tabular}{lrcccc}
\hline
\textbf{Method} & \textbf{N} & \textbf{Both} & \textbf{Semantic Only} & \textbf{Realism Only} & \textbf{Neither} \\
\hline
\multicolumn{6}{l}{\textit{Rule-based Methods}} \\
imgaug & 1059 & 1.7\% & 0.0\% & 97.5\% & 0.8\% \\
albumentations & 1092 & 16.6\% & 0.7\% & 82.1\% & 0.5\% \\
\hline
\multicolumn{6}{l}{\textit{Generative AI methods}} \\
OpenAI & 204 & 2.9\% & 74.0\% & 17.6\% & 5.4\% \\
Qwen & 75 & 1.3\% & 66.7\% & 28.0\% & 4.0\% \\
Gemini & 141 & 2.1\% & 36.2\% & 58.2\% & 3.5\% \\
Flux & 537 & 19.0\% & 1.3\% & 77.7\% & 2.0\% \\
\hline
\end{tabular}
\end{table}

Rule-based Methods fail almost exclusively due to unrealistic appearance while preserving semantic content. imgaug shows 0.0\% semantic-only failures, indicating it never alters scene content---when rejected, the sole issue is artificial appearance. In contrast, top-performing generative methods (OpenAI, Qwen) show the inverse pattern: when they fail, it is predominantly due to semantic alterations rather than unrealistic appearance. Flux represents an intermediate case, exhibiting failure patterns closer to Rule-based Methods, which aligns with its lower VLM acceptance rates reported in Section~4.

\subsection{Failure Modes by Target Condition}

Table~\ref{tab:failure_by_condition} presents failure patterns across target environmental conditions.

\begin{table}[htbp]
\centering
\caption{Failure type distribution by target condition. Percentages are row-wise.}
\label{tab:failure_by_condition}
\begin{tabular}{lrcccc}
\hline
\textbf{Condition} & \textbf{N} & \textbf{Both} & \textbf{Semantic Only} & \textbf{Realism Only} & \textbf{Neither} \\
\hline
Fog & 426 & 0.0\% & 10.1\% & 88.3\% & 1.6\% \\
Rain & 849 & 4.8\% & 8.8\% & 85.5\% & 0.8\% \\
Night & 999 & 14.1\% & 4.6\% & 79.9\% & 1.4\% \\
Snow & 834 & 15.5\% & 12.4\% & 70.3\% & 1.9\% \\
\hline
\end{tabular}
\end{table}

Fog augmentation failures are almost entirely realism-only (88.3\%), with no cases of combined failures. This aligns with fog being the easiest condition to simulate: methods either achieve realistic fog or produce obviously artificial haze, but rarely alter semantic content in the process. Snow shows the highest rate of combined failures (15.5\%) and semantic-only failures (12.4\%), suggesting this condition is particularly prone to unintended content modifications. Night augmentation exhibits elevated combined failure rates (14.1\%), indicating that the global illumination changes required for day-to-night transformation can simultaneously produce unrealistic lighting and alter scene semantics.

\subsection{Method-Condition Interactions}

Table~\ref{tab:failure_method_condition} presents failure counts for selected method-condition combinations that reveal notable patterns.

\begin{table}[htbp]
\centering
\caption{Failure type distribution for selected method-condition combinations with highest failure counts or notable patterns.}
\label{tab:failure_method_condition}
\begin{tabular}{llrcccc}
\hline
\textbf{Method} & \textbf{Condition} & \textbf{N} & \textbf{Both} & \textbf{Sem.} & \textbf{Real.} & \textbf{Neither} \\
\hline
albumentations & snow & 348 & 32.5\% & 0.3\% & 66.4\% & 0.9\% \\
flux & night & 294 & 34.7\% & 0.3\% & 63.3\% & 1.7\% \\
imgaug & snow & 330 & 3.9\% & 0.0\% & 94.8\% & 1.2\% \\
imgaug & night & 303 & 0.7\% & 0.0\% & 98.7\% & 0.7\% \\
openai & rain & 66 & 0.0\% & 93.9\% & 0.0\% & 6.1\% \\
qwen & snow & 27 & 0.0\% & 96.3\% & 0.0\% & 3.7\% \\
qwen & fog & 9 & 0.0\% & 100.0\% & 0.0\% & 0.0\% \\
\hline
\end{tabular}
\end{table}

Several patterns warrant attention. The Flux/night combination shows the highest ``both'' failure rate (34.7\%) of any method-condition pair, indicating that Flux's nighttime augmentations frequently produce both unrealistic lighting and semantic alterations. This aligns with its low VLM acceptance (0.176) reported in Section~4. In contrast, when OpenAI's rain augmentations fail, the failures are almost exclusively semantic (93.9\%)---the rain effects appear realistic, but scene content is modified. Similarly, Qwen's snow and fog failures are entirely semantic (96.3\% and 100.0\%), suggesting these methods have largely solved the realism challenge but struggle to maintain perfect semantic fidelity.

\subsection{Inter-Judge Agreement}

The three LLM judges showed strong agreement in classifying failure reasons (Table~\ref{tab:llm_agreement}).

\begin{table}[htbp]
\centering
\caption{Inter-judge agreement for failure classification.}
\label{tab:llm_agreement}
\begin{tabular}{lccc}
\hline
\textbf{Classification} & \textbf{Fleiss' $\kappa$} & \textbf{Unanimous (3/3)} & \textbf{Majority (2/3)} \\
\hline
Realism & 0.904 (almost perfect) & 97.4\% & 2.6\% \\
Semantic & 0.749 (substantial) & 88.6\% & 11.4\% \\
\hline
\end{tabular}
\end{table}

The higher agreement on realism classification ($\kappa = 0.904$) compared to semantic classification ($\kappa = 0.749$) suggests that unrealistic appearance is more objectively identifiable from textual descriptions than semantic changes. Table~\ref{tab:llm_judge_variation} shows the classification rates by individual LLM judge.

\begin{table}[htbp]
\centering
\caption{Classification rates by LLM judge, showing inter-judge variation.}
\label{tab:llm_judge_variation}
\begin{tabular}{lcc}
\hline
\textbf{LLM Judge} & \textbf{Semantic Failure Rate} & \textbf{Realism Failure Rate} \\
\hline
Gemini & 22.2\% & 90.8\% \\
GPT & 19.1\% & 88.9\% \\
Claude & 14.5\% & 90.3\% \\
\hline
\textit{Spread (max--min)} & 7.7\% & 1.9\% \\
\hline
\end{tabular}
\end{table}

The 4$\times$ difference in inter-judge spread (7.7\% for semantic vs.\ 1.9\% for realism) indicates that unrealistic appearance is easier to recognize from VLM textual descriptions. When a VLM states that ``the fog looks like a filter'' or ``the snow doesn't accumulate properly,'' all judges agree it constitutes a realism issue. However, when descriptions mention ``objects are obscured'' or ``details are lost,'' judges show more uncertainty about whether this constitutes a semantic change versus an expected consequence of the target condition.

\subsection{Illustrative Examples}

To illustrate the failure categories, we provide representative examples from the VLM judge explanations.

\textbf{Realism-only failure} (albumentations fog): ``The semantic content of the image is preserved; all the objects from the original image, like the buildings, trees, and tram tracks, are present in the augmented version. However, the fog simulation is not realistic. The fog is applied as a uniform, semi-transparent white layer over the entire image. Real fog exhibits depth-dependent density [...] the foreground elements are faded to a similar degree as the distant buildings, which makes the effect look like a simple filter rather than a natural atmospheric condition.''

\textbf{Semantic-only failure} (Qwen fog): ``The added fog is realistic in its appearance, with convincing density that increases with distance and a natural-looking diffusion of light from the traffic signals. However, the semantic preservation criterion is not met. Key objects from the original scene, specifically the black car on the left and the white van in the center of the road, have been completely removed in the augmented image, not just obscured by the fog.''

\textbf{Both failure types} (albumentations snow): ``The augmented image fails on both criteria. For semantic preservation, the image is severely overexposed and washed out, losing almost all color and detail from the original. [...] For snow realism, the effect is not convincing at all. There is no visible snow on the ground, trees, or buildings, nor is there any falling snow. The image simply looks like a heavily overexposed or foggy version of the original, not a snowy scene.''

\subsection{Limitations}

This analysis has several limitations. First, we classify failure reasons post-hoc by analyzing VLM textual explanations rather than directly evaluating semantic preservation and realism as separate criteria. A cleaner design would deploy separate VLM judges for each criterion. Second, the ``neither'' category (1.4\%) represents cases where the VLM's explanation was insufficiently specific, highlighting the dependence on VLM articulation quality. Third, the classification inherently depends on how VLM judges express their reasoning; different judges may emphasize different aspects of the same underlying issue.

\subsection{Summary}

The failure mode analysis reveals a fundamental trade-off between augmentation paradigms. Rule-based Methods preserve semantic content perfectly but produce obviously artificial effects. Generative methods achieve visual realism but at the cost of occasional semantic alterations. The high inter-judge agreement ($\kappa = 0.749$--$0.904$) validates the reliability of these classifications. These findings suggest that practitioners must weigh the relative importance of semantic fidelity versus visual realism when selecting augmentation methods for their specific application requirements.

\clearpage
\section{Metric Divergence Analysis}
\label{app:metric_divergence}

A natural concern is whether disagreement between VLM jury acceptance and embedding-based Mahalanobis distance indicates that one metric is simply less reliable than the other. We conducted targeted qualitative analysis to investigate which metric better captures meaningful realism when they disagree.

\subsection{Methodology}

We identified cases of maximal disagreement between metrics by selecting:
\begin{enumerate}
    \item \textbf{VLM-accepted, high Mahalanobis}: Images unanimously accepted by all three VLM judges (indicating perceptual realism) but with the highest Mahalanobis distances (indicating statistical dissimilarity to real adverse-condition distributions).
    \item \textbf{VLM-rejected, low Mahalanobis}: Images unanimously rejected by all three VLM judges but with the lowest Mahalanobis distances (indicating statistical similarity to real distributions).
\end{enumerate}

\begin{figure}[htbp]
    \centering
    \includegraphics[width=0.95\linewidth]{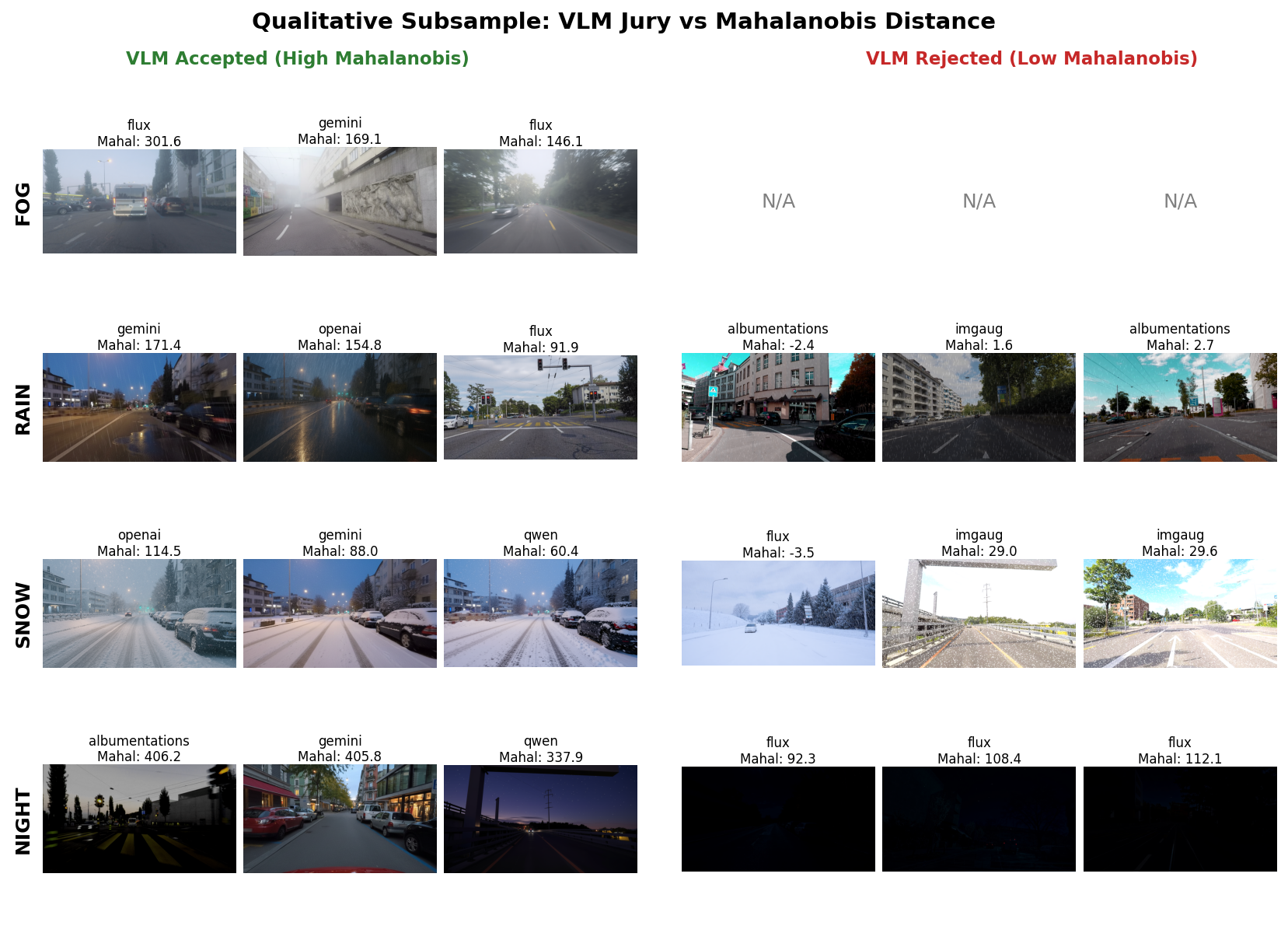}
    \caption{Cases of maximal disagreement between VLM jury and Mahalanobis distance. Left (green): images unanimously accepted by VLMs despite high Mahalanobis distance. Right (red): images unanimously rejected by VLMs despite low Mahalanobis distance. Visual inspection suggests VLM judgments better capture perceptual realism.}
    \label{fig:metric_divergence}
\end{figure}

From 1,025 augmented images, 410 (40.0\%) achieved unanimous VLM acceptance and 187 (18.2\%) achieved unanimous rejection, with 428 (41.8\%) showing mixed votes. We selected the top examples from each disagreement category per condition for qualitative inspection. Figure~\ref{fig:metric_divergence} presents representative cases.

\subsection{Qualitative Findings}

Visual inspection reveals systematic patterns favouring VLM judgments as the more reliable indicator of perceptual realism.

\textbf{Rain condition}: VLM-rejected images with low Mahalanobis distance (e.g., albumentations at $d_{\mathrm{rel}}=-2.4$, imgaug at $d_{\mathrm{rel}}=1.6$) exhibit colour shifts toward cyan/blue tones without visible precipitation or wet surfaces. These images are statistically proximate to real rain imagery---likely because the colour temperature shift resembles overcast conditions---yet lack the defining visual characteristics of rain. Conversely, VLM-accepted images from generative methods show convincing rain streaks and surface reflections despite higher Mahalanobis distances.

\textbf{Snow condition}: The lowest-distance rejected image (Flux, $d_{\mathrm{rel}}=-3.5$) shows a washed-out scene lacking visible snow accumulation or falling precipitation. Rule-based rejections (imgaug at $d_{\mathrm{rel}}\approx29$--$30$) display grainy noise overlays that statistically resemble snow texture but appear obviously artificial. VLM-accepted generative augmentations show coherent snow accumulation on surfaces despite greater statistical distance.

\textbf{Night condition}: This condition shows the most dramatic divergence. VLM-rejected Flux augmentations ($d_{\mathrm{rel}}=92$--$142$) render scenes nearly or completely black, obscuring all semantic content. These achieve relatively low Mahalanobis distances because extreme darkness is statistically similar to real nighttime imagery in embedding space. However, VLM judges correctly identify that useful nighttime driving imagery requires visible scene content with appropriate artificial lighting---a criterion that pure distributional similarity cannot capture. VLM-accepted nighttime augmentations from Gemini and Qwen ($d_{\mathrm{rel}}=337$--$406$) maintain scene visibility with realistic streetlights and headlight effects despite being statistically distant from the reference distribution.

\textbf{Fog condition}: Notably, no fog augmentations achieved unanimous VLM rejection, preventing direct comparison. This aligns with fog being the easiest condition to simulate, where even simple contrast reduction approximates the target appearance.

\subsection{Interpretation}

These observations suggest that when VLM judgments and Mahalanobis distances disagree, VLM judgments more reliably indicate perceptual realism. Several factors explain this pattern:

First, Mahalanobis distance measures statistical proximity to a reference distribution but cannot distinguish between ``realistic but rare'' and ``unrealistic.'' An image that is statistically similar to real adverse-condition imagery may nonetheless lack the perceptual cues that define the condition (e.g., colour-shifted images resembling overcast weather without actual rain).

Second, VLMs encode images into high-dimensional representations during processing, yet their judgments diverge from pure distributional similarity. This suggests they apply learned perceptual criteria---likely acquired from training on image-caption pairs describing weather conditions---that go beyond statistical proximity.

Third, the nighttime condition illustrates a fundamental limitation of embedding-based evaluation: completely dark images are statistically similar to real night imagery but fail the functional requirement that scene content remain visible. VLM judges, prompted with explicit criteria about ``appropriate darkness'' and ``objects remaining visible,'' correctly penalise such failures.

\subsection{Limitations}

This analysis has limitations. We examined only cases of unanimous VLM agreement, excluding the 41.8\% of images with mixed votes where metric comparison is less clear-cut. The qualitative assessment relies on our visual inspection rather than independent human annotators. Additionally, the reference distributions were fitted on ACDC imagery; embedding-based metrics might perform differently with alternative reference datasets.

\subsection{Conclusion}

While we present both evaluation modalities precisely because neither constitutes a gold standard, this analysis suggests that VLM judgments provide a more reliable indicator of perceptual realism when metrics disagree. Embedding-based analysis remains valuable for detecting distributional shifts and providing an independent validation signal, but practitioners should weight VLM acceptance more heavily when the two metrics conflict.

\clearpage
\section{Augmentation Prompts}
\label{app:prompts}

Table~\ref{tab:prompts} presents the natural language prompts used for each environmental condition across all generative AI methods. All models received identical prompts to ensure fair comparison. Each prompt consists of a task description followed by specific criteria.

\begin{table}[htbp]
\centering
\caption{Prompts used for generative AI augmentation methods.}
\label{tab:prompts}
\begin{tabular}{lp{10cm}}
\hline
\textbf{Condition} & \textbf{Prompt} \\
\hline
Fog & \textbf{Task:} In the image changes the weather so it is a foggy day. \newline
\textbf{Criteria:} \newline
$\bullet$ The image should be identical in every way to the provided image except that it should be foggy. \newline
$\bullet$ The camera angle and perspective must be the same. \newline
$\bullet$ The fog should be light and volumetric, getting denser with distance. \\
\hline
Rain & \textbf{Task:} In the image changes the weather so it is a rainy day. \newline
\textbf{Criteria:} \newline
$\bullet$ The image should be identical in every way to the provided image except that it should be rainy. \newline
$\bullet$ The camera angle and perspective must be the same. \newline
$\bullet$ The rain should be visible with natural rain streaks and wet surfaces. \\
\hline
Snow & \textbf{Task:} In the image changes the weather so it is a snowy day. \newline
\textbf{Criteria:} \newline
$\bullet$ The image should be identical in every way to the provided image except that it should be snowy. \newline
$\bullet$ The camera angle and perspective must be the same. \newline
$\bullet$ The snow should be falling and accumulated on horizontal surfaces. \\
\hline
Night & \textbf{Task:} In the image changes the time of day so it is nighttime. \newline
\textbf{Criteria:} \newline
$\bullet$ The image should be identical in every way to the provided image except that it should be nighttime. \newline
$\bullet$ The camera angle and perspective must be the same. \newline
$\bullet$ The lighting should be darker with evening/twilight atmosphere but objects should remain clearly visible. \\
\hline
\end{tabular}
\end{table}

\end{document}